\documentclass[review]{elsarticle}

\usepackage{lineno}
\modulolinenumbers[5]
\usepackage{balance} 
\usepackage{natbib}
\usepackage{wrapfig}
\usepackage[utf8]{inputenc} 
\usepackage[T1]{fontenc}    
\usepackage{hyperref}       
\usepackage{url}            
\usepackage{booktabs}       
\usepackage{amsfonts}       
\usepackage{nicefrac}       
\usepackage{microtype}      
\usepackage{amsmath}
\usepackage{algorithm}
\usepackage{algpseudocode}
\usepackage{graphicx}
\usepackage{caption}
\usepackage{subcaption}
\usepackage{xcolor}

\usepackage{xparse}
\usepackage{blindtext}
\usepackage{morewrites}
\usepackage{xkeyval}
\usepackage{tcolorbox}
\newwrite\authorbibfile
\AtBeginDocument{%
  \immediate\openout\authorbibfile=\jobname.aub%
}%
\AtEndDocument{%
\immediate\closeout\authorbibfile
\InputIfFileExists{\jobname.aub}{}{}
}%

\makeatletter
\define@key{authorbib}{scale}[1]{%
\def\AuthorbibKVMacroScale{#1}%
}
\define@key{authorbib}{wraplines}[10]{%
\def\AuthorbibKVMacroWraplines{#1}%
}
\define@key{authorbib}{imagewidth}[4cm]{%
\def\AuthorbibKVMacroImagewidth{#1}%
}
\define@key{authorbib}{overhang}[10pt]{%
\def\AuthorbibKVMacroOverhang{#1}%
}
\define@key{authorbib}{imagepos}[l]{%
\def\AuthorbibKVMacroImagepos{#1}%
}
\makeatother

\presetkeys{authorbib}{imagepos=l, imagewidth=4cm, wraplines=8, overhang=20pt}{}
\newlength{\AuthorbibTopSkip}
\newlength{\AuthorbibBottomSkip}
\NewDocumentCommand{\authorbibliography}{+o+m+m+m}{%
  \IfNoValueTF{#1}{%
  }{%
    \setkeys{authorbib}{#1}%
    \immediate\write\authorbibfile{%
      \string\begin{wrapfigure}[\AuthorbibKVMacroWraplines]{\AuthorbibKVMacroImagepos}[\AuthorbibKVMacroOverhang]{\AuthorbibKVMacroImagewidth}^^J
        \string\includegraphics[scale=\AuthorbibKVMacroScale]{#2}^^J
        \string\end{wrapfigure}^^J
    }%
  }%
  \IfNoValueTF{#3}{%
    \typeout{Warning: No author name}%
  }{%
      \immediate\write\authorbibfile{%
      \unexpanded{\vspace{\AuthorbibTopSkip}}^^J
      \string\noindent\relax
      \unexpanded{\textbf{#3}\par}^^J
      \string\noindent\relax
      \unexpanded{#4}^^J%
      \unexpanded{\vspace{\AuthorbibBottomSkip}}^^J
      }%
  }%
}%

\journal{Information Sciences}

\begin{document}
\begin{frontmatter}

\title{Reward Shaping Using Convolutional Neural Network}


\author[Concordia]{Hani Sami}
\author[khalifa]{Hadi Otrok}
\author[Concordia,khalifa]{Jamal Bentahar}
\author[lau,nyu]{Azzam Mourad}
\address[Concordia]{Concordia Institute for Information Systems Engineering, Concordia University, Montreal, Canada}
\address[khalifa]{Center of Cyber-Physical Systems (C2PS), Department of EECS, Khalifa University, Abu Dhabi, UAE}
\address[lau]{Department of Computer Science and Mathematics, Lebanese American University, Beirut, Lebanon}
\address[nyu]{Division of Science, New York University, Abu Dhabi, UAE}
\author[khalifa]{Ernesto Damiani\corref{mycorrespondingauthor}}

\cortext[mycorrespondingauthor]{Corresponding author}

    


\begin{abstract}
In this paper, we propose Value Iteration Network for Reward Shaping (VIN-RS), a potential-based reward shaping mechanism using Convolutional Neural Network (CNN). The proposed VIN-RS embeds a CNN trained on computed labels using the message passing mechanism of the Hidden Markov Model. The CNN processes images or graphs of the environment to predict the shaping values. Recent work on reward shaping still has limitations towards training on a representation of the Markov Decision Process (MDP) and building an estimate of the transition matrix. The advantage of VIN-RS is to construct an effective potential function from an estimated MDP while automatically inferring the environment transition matrix. The proposed VIN-RS estimates the transition matrix through a self-learned convolution filter while extracting environment details from the input frames or sampled graphs. Due to (1) the previous success of using message passing for reward shaping; and (2) the CNN planning behavior, we use these messages to train the CNN of VIN-RS. Experiments are performed on tabular games, Atari 2600 and MuJoCo, for discrete and continuous action space. Our results illustrate promising improvements in the learning speed and maximum cumulative reward compared to the state-of-the-art.
\end{abstract}

\begin{keyword}
Reinforcement Learning, Reward Shaping, CNN, VIN, Message Passing, Atari, MuJoCo
\end{keyword}
\end{frontmatter}

\section{Introduction}

A Reinforcement learning (RL) algorithm is executed against a Markov Decision Process (MDP) environment. The MDP environment is sketched by the solution designer where the agent can perform actions decided by RL. The agent sends feedback to the RL solution with a reward value to update the value function based on the actions taken. Therefore, an accurate representation of the reward function is vital for future action selection. Hence, dynamicity in the structure of the reward function is required to adapt to environmental changes and produce more effective rewards. Consequently, better rewards lead to faster convergence to near optimality with regard to agent decisions. MDP environments have different types with continuous or discrete, finite or infinite states, and action spaces. Furthermore, a transition matrix deciding the next state of the agent is most of the time unknown. These MDP properties make it challenging to develop a scalable and dynamic reward function \cite{oh2020discovering}.

RL algorithms are slow to converge, where most of the time is spent on exploration at the early stages of learning. There are multiple learning speedup techniques for RL such as offline learning, dynamic exploration, transfer learning, imitation learning, and reward shaping \cite{shurrab2022iot, alagha2022target}. Reward shaping alters the original reward function with values generated from a shaping function. The shaping values speed learning in RL by directing the reward function to speed the policy convergence \cite{ng1999policy}.  One of the reward shaping mechanisms is potential-based, which ensures that updates to the original reward function do not affect the ability of an agent to reach optimal policy decisions. Due to the different types of MDP and dynamicity of the environment, it is difficult to design a scalable and effective potential function for reward shaping that is suitable for most environments \cite{AmodeiOSCSM16}. The existing reward shaping solutions suffer from one or more of the following limitations: (1) they alter the optimal policy; (2) they are based on action exploration only \cite{garaffa2021reinforcement}; (3) they are not applicable in different environments; (4) they rely on transition matrix approximation; (5) have limited representation of sampled MDPs. The negative consequence of the first limitation is that obtaining the optimal policy cannot be guaranteed. Because of the second limitation, the reward function is never adapted to the environment. Due to the other limitations, the agent performance based on the potential function can be improved. The applicability of a reward shaping solution is measured by the performance in various environments and the dependability on external knowledge like expert feedback. Therefore, a potential-based reward shaping solution is still an open problem. \textit{Hence, we propose in this paper a novel potential-based reward shaping solution that is scalable, learns on a representation of the MDP either through frames of images or sampled MDP graphs, and estimates a transition matrix while training.}

The proposed potential function architecture follows the mechanism of the Value Iteration Network (VIN) and uses convolution layers to perform planning \cite{tamar2016value}. The original VIN module in \cite{tamar2016value} uses a Convolutional Neural Network (CNN) architecture, which can be trained using RL or Imitation Learning (IL). The CNN of the VIN can perform value iteration on an MDP for planning. The output of the CNN is part of an attention mechanism that selects actions as part of the optimal plan. In \cite{tamar2016value}, learning is either done using IL, thus requiring a large number of labels, or using RL whose performance is poor on irregular graphs. Irregular graphs are problematic when training VIN with RL because the number of actions for the neighbors varies.

The proposed VIN-RS trains a novel CNN based on the probabilistic view of RL to serve as a potential function. Our new scheme, named Value Iteration Network for Reward Shaping (VIN-RS), incorporates this novel training mode and new architecture tailored for reward shaping. Computing message passing on Hidden Markov Models (HMM) \cite{toussaint2006probabilistic}, composed of forward and backward messages, derives the probability of trajectory optimality, thus can be used to redirect the reward function and speed learning \cite{toussaint2006probabilistic, klissarov2020reward}. Computing these messages in a large environment requires high computation as discussed further in Section \ref{Message_Passing}. On one hand, message passing helps the agent decide if the current state belongs to an optimal trajectory. On the other hand, the CNN module of VIN provides a goal-oriented plan for an optimal trajectory. In this paper, we will argue that the message passing and VIN combined would help accelerate learning by acting as the potential function. Therefore, we aim to train the proposed CNN of VIN-RS based on the message passing loss computed from sampled trajectories followed by the agent.

In\cite{klissarov2020reward}, the authors propose a potential-based reward shaping solution using Graph Convolution Network (GCN). GCN has the potential in propagating the forward and backward messages of HMM using the graph operation for sharing information between nodes. Despite significant improvement achieved over existing reward shaping solutions, using GCN as the potential function still has some limitations, and it cannot generalize to all environments. Even though both CNN and GCN belong to the family of Graph Neural Networks (GNN), GCN performs message passing on a sample of the states, while CNN uses full images of the environment, which can reveal more states and speed planning. In addition, to perform the GCN layer operation, the transition matrix of the MDP should be estimated \cite{klissarov2020reward}. Assuming that the value function is smooth over the MDP graph, the graph Laplacian is used as an approximation of the transition matrix, resulting in a margin of error \cite{petrik2007analysis}. Furthermore, GCN has issues related to MDP representation and information extraction, which is due to the graph approximation technique used to represent the environment. Compared to GCN, the proposed CNN learns a representation of this transition matrix while training. More specifically, VIN-RS estimates a new MDP, not necessarily related to the original one, which is learned using images of the environment and trained following the message passing values. Even when the environment is not captured as images, VIN-RS can train the CNN on a graph representation from this environment, which leads to promising performance as illustrated later in our experiments.

Training VIN-RS for reward shaping requires the true labels that are inferred using message passing. Therefore, the loss function incorporates the message passing mechanism. Due to (1) the computation complexity induced by calculating the message passing; and (2) the difficulty of retrieving and sometimes approximating the transition matrix of the environment, we train CNN on image frames or samples of the environment graph capturing the current state of the agent. CNN uses a transition matrix that is trained as part of the network. The output of the proposed CNN is a regression value for each possible action in the environment that estimates if the current state of the agent belongs to an optimal trajectory. Furthermore, VIN-RS embeds the look-ahead advice mechanism \cite{wiewiora2003principled}. By evaluating the state and action, the look-ahead advice improves the quality of the potential function and thus can improve the learning speed using reward shaping. Thus, VIN-RS produces advice at the state and action levels, which is at the core of the look-ahead advice mechanism. In addition, it is possible to apply our solution in discrete and continuous action spaces.


The contributions of this paper are summarized as follows:

\begin{enumerate}
    \item VIN-RS for potential-based reward shaping using a novel CNN architecture. 
    \item CNN trained on images or graphs from the environment while using a message passing mechanism in the loss function.
    \item Estimating the transition matrix through training the CNN.
    \item Overcoming the limitations of using GCN for reward shaping as the potential function by training on an estimated MDP using CNN.
    \item State-of-the-art results in most of the games for discrete and continuous action spaces.
\end{enumerate}

The structure of the paper is divided as follows. In We describe the related work in Section \ref{sec:related_work}. In Section \ref{sec:prelimiaries}, we present the background required to cover the core concepts of our approach. In Section \ref{sec:proposed_scheme}, we present our proposed reward shaping solution "VIN-RS". The evaluations conducted on tabular, Atari and MuJoCo games are illustrated in Section \ref{sec:experiments}. The results of our evaluation shows that VIN-RS achieves on average the best results in the Atari and MuJoCo games compared to all baselines. Finally, we summarize and conclude the paper in Section \ref{sec:conclusion} with future directions.

\section{Related Work}
\label{sec:related_work}

Reward shaping has gained more attention in the past years due to the importance of speeding the learning process in Deep RL \cite{mnih2013playing}, especially in applications requiring real-time feedback. In this section, we first describe VIN and its applications. Second, we discuss existing reward shaping solutions and their limitations. Third, we discuss and present the limitations of recent literature solutions that utilize deep learning to build the potential function.

\subsection{Value Iteration Networks and Applications}
In \cite{tamar2016value}, a VIN module is proposed to perform planning on a new MDP extracted from images of the environment. The motivation behind VIN is to support planning in NNs by integrating a new value iteration method inspired by the policy updates in RL. The value of the work comes from using the convolution operators to perform value iteration on an unknown MDP that outputs an optimal trajectory. The states and actions of this MDP are the same as the original MDP, while the reward and transition functions are any differentiable functions that can be learned from the model while training. One of the main limitations of VIN is that it can be applied using imitation learning which requires a lot of ground truth labels or RL that provides poor performance. Furthermore, VIN supports low-dimensional environments or MDPs only. Due to the importance of the planning feature in NNs, various works extended form \cite{tamar2016value} and proposed new differentiable reward and transition functions that can stabilize learning in the network. For example, the authors in \cite{niu2018generalized} proposes a novel convolution operator to learn and plan on spatial and irregular graphs.

Various applications use VIN to combine planning with a supervised learning task to improve the quality of the decisions. For instance, the work in \cite{li2021dynamic} utilizes VIN to perform UAV planning and adapt to novel physical locations. Furthermore, the work in \cite{yang2018learning} uses VIN to learn urban navigation planning.

In this work, we propose a new variation of VIN, named VIN-RS, for the first time in the context of potential-based reward shaping. VIN-RS encodes a new training mode with message passing as part of the loss function. Our VIN-RS can effectively plan trajectories to speed learning in the original policy. VIN-RS takes images of the environment as input to output shaping values. The shaping values are used in the form of potential-based reward shaping by appending the original reward function.

\subsection{Existing Reward Shaping Methods}
There exist various reward shaping solutions that (1) alter the optimal policy, (2) target action exploration instead of reward signal, or (3) requires human intervention. For instance, Learning Intrinsic Rewards for Policy Gradients (LIRPG) proposes an optimal reward framework and does not guarantee invariance of the optimal policy \cite{zheng2018learning}. The Random Network Distillation (RND) approach provides an action exploration method to accelerate learning \cite{burda2018exploration}. RND uses an exploration bonus, which is calculated using the error of a NN that predicts the observation features. The proposed RND solution provides Superior performance in the challenging Montezuma Revenge game. In \cite{pathak2017curiosity}, authors propose the intrinsic curiosity module (ICM) to speed learning using action exploration. In ICM, the authors formulate exploration by the error of the agent when trying to predict the consequences of its actions \cite{pathak2017curiosity}.
In contrast to RND and ICM, our proposed VIN-RS offers reward shaping based on the reward signal and not the exploration bonus. 
In \cite{klissarov2020reward}, the authors proposed a potential-based reward shaping solution that performs message passing using GCN. More details about the GCN solution are provided in the next subsection.
In Section \ref{sec:experiments}, we compare the performance of VIN-RS to LIRPG \cite{zheng2018learning}, RND \cite{burda2018exploration}, ICM \cite{pathak2017curiosity}, and GCN \cite{klissarov2020reward}.

\subsection{Using GCN for Reward Shaping}
\label{subs:related_gcn}

Existing reward shaping solutions requires a neural network to become scalable and accomodate for dynamicity in the environment with large state or action spaces. For instance, the work in \cite{grzes2010online} suffers from scalability issues, while \cite{harutyunyan2015shaping} demands human intervention to update the reward function with feedbacks. Therefore, we study the related work proposing to build the potential function through deep learning to overcome the mentioned problems. In specific, the most suitable deep learning models belong to the family of Graph Neural Network (GNN), such as GCN. Hence, we discuss the two most related work that uses GCN to build the value function \cite{klissarov2020reward, sami2022graph}. The GCN is capable of recursively propagate messages among neighboring nodes in the graph, determining its relation and importance. When using message passing of HMM, those messages reveal more information about the state or trajectory of state optimality, thus enlightening the original reward function about useful information among the selected path. In \cite{klissarov2020reward}, an improvement of learning speed and reward achieved are presented for the first time when using GCN as a reward shaping function. Despite its performance, the presented mechanism suffers from various limitations including the representation of the sampled MDP as a sub-graph of sampled transitions, and the approximation of the transition matrix using graph Laplacian, which results in a margin of error affecting its performance.

In this paper, we replace GCN with a VIN-RS that utilizes a CNN to perform reward shaping as a potential function. In addition, GCN utilizes a subset of the states and forms a graph that is passed to GCN for training. However, using VIN-RS, we pass images of the environment as input to train the potential function; therefore, more knowledge is provided with a larger set of states. Furthermore, using GCN to compute the message passing requires an approximation of the transition matrix. In this case, the graph Laplacian is utilized \cite{klissarov2020reward}. Due to many drawbacks of using graph Laplacian as analyzed in \cite{petrik2007analysis}, reward shaping using GCN affects the performance.
Several methods were proposed to construct bases for Value Function Approximation (VFA), such as using the graph Laplacian. It is proven that the graph Laplacian can only produce effective VFA when assuming that the latter is smooth over the induced MDP graph. Therefore, using the graph Laplacian to approximate the transition matrix cannot generalize to all MDPs \cite{petrik2007analysis}. In our previous work \cite{sami2022graph}, the graph Laplacian is replaced by a Krylov subspace computed using the augmented Krylov \cite{petrik2007analysis}, as an attempt to overcome the graph Laplacian limitations. However, this method still cannot guarantee an improvement over GCN in some cases. Therefore, we replace in this work GCN by a CNN. In the proposed VIN-RS, the probability transition matrix within the convolution layers is learned when training the CNN network, thus avoiding the burden of approximating this matrix.

\section{Background}
\label{sec:prelimiaries}
A modeled MDP environment $\mathcal{M}$ requires a clear definition of the state space $\mathcal{S}$, action space $\mathcal{A}$, and reward function $r$. As part of the MDP, $\gamma$ is also predefined to a value close to one indicating the amount of information to propagate during the value function updates in cumulative future steps. Furthermore, a transition matrix $P$ is defined to foster the transitions from previous state to next state following an action decided through exploration or exploitation. Thus, an MDP is represented by the following tuple $(\mathcal{S}, \mathcal{A}, P, r, \gamma)$. The policy $\pi$ is formed through deterministic or stochastic updates based on the environment feedback from the environmnet to maximize the cumulative future discounted rewards. In this work, we focus on value function updates calculated as follows: $\mathcal{V}_{n+1}(s) = \text{max}_a \mathcal{Q}_n(s, a)$, where $\mathcal{Q}_n(s, a) = r(s, a) + \gamma \sum_{s'}\mathcal{P}(s'|s, a)\mathcal{V}_n(s')$, $s\in\mathcal{S}$ and $a\in\mathcal{A}$. An optimal value function $\mathcal{V}^*(s)$ is the maximal return obtained over steps for all trajectories. When $n\rightarrow\infty$, the value function $V_n$ converges. Thus an optimal policy $\pi^*$ is obtained as follows: $\pi^*(s)=\text{argmax}_a Q_{\infty}(s, a).$

\subsection{Reward Shaping}
\label{reward_shaping}
A reward shaping is the process of updating the reward value by a shaping value produced using a shaping function. This shaping value is computed by passing the current and next states to the shaping function. We denote the shaping function as $F$ and the updated reward function takes the following form:
\begin{equation}
    R(s, a, s') = r(s, a) + F(s, s')
\end{equation}
where $F(s, s')$ is the shaping function. In terms of potential-based reward shaping, the shaping function is computed as follows  \cite{wiewiora2003principled}:

\begin{equation}
    F(s, s') = \gamma\phi(s') - \phi(s)
    \label{eq:reward_shaping}
\end{equation}
where $\phi$ is a potential-shaping function. 
More explanation of previous approaches proposing shaping functions is described in Section \ref{sec:related_work}.

\subsection{Convolutional Neural Network}
The CNN network is proven to be successful in computer vision and natural language processing applications. Furthermore, CNN layers extract features from the images as it goes deeper into the network. The input layer in a CNN takes as input raw image pixels which are three-dimensional. A CNN is composed of convolutional layer, non-linearity layer, max pooling layer, and fully connected layer. At each convolutional (conv) layer, there is a kernel matrix or filter with a certain size. Setting the filter size is considered a hyperparameter. A filter for the current layer is applied on the input matrix, where matrix multiplication takes place. In addition, a sliding window or stride on the input matrix can be applied so that more information about the image can be concluded. Adding more layers with various filters results, in most cases, in more feature extraction from the input matrix. Furthermore, padding can be added to the images to not lose any information present at the frame or edges of the matrix.

\subsection{Value Iteration Network}
A VIN incorporates planning inside a policy of the original MDP $\mathcal{M}$ by performing value iteration using a CNN \cite{tamar2016value}. It is assumed that VIN tries to learn and solve another MDP $\bar{\mathcal{M}}$. Similar to any MDP, $\bar{\mathcal{M}}$ has states, actions, reward and transition functions denoted as $\bar{{s}}\in\bar{\mathcal{S}}, \bar{{a}}\in\bar{\mathcal{A}}, \bar{\mathcal{R}}(\bar{{s}}, \bar{{a}}), \text{and } \bar{\mathcal{P}}(\bar{{s'}}|\bar{{s}}, \bar{{a}})$ respectively. The state and action spaces in $\bar{\mathcal{M}}$ are similar to $\mathcal{M}$. The reward and transition matrices of $\bar{\mathcal{M}}$ depend on the observations of $\mathcal{M}$, i.e. $\bar{\mathcal{R}} = f_r(\phi(s))$ and $\bar{\mathcal{P}} = f_p(\phi(s))$. The functions of the reward and transition ($f_r$ and $f_p$) are learned during the policy training of $\bar{\mathcal{M}}$ using the CNN. The learned policy of $\bar{\mathcal{M}}$ is connected to obtaining the optimal policy of $\mathcal{M}$, even though the reward and transition functions are not the same. The input to the VIN model is a list of images extracted from the environment. The first layer in the CNN of VIN is a convolution (Conv) layer that processes raw pixels and pass them to the second layer. The first step in value iteration at the second conv layer is to convert the input of the previous layer into a reward matrix using the reward function $f_r$. The filter/kernel of this second conv layer is considered as the probability transition matrix of $\bar{\mathcal{M}}$. In addition, the third conv layer contains the $\mathcal{Q}$ value or state-action value function over the channels of this layer for $\bar{\mathcal{M}}$. Finally, this last layer is max-pooled to produce the next value iteration $\bar{\mathcal{V}}$.

The output of VIN is only for a subset of states. Therefore, the output is passed to an attention module that helps reduce the number of parameters to train or the actions to focus on. Furthermore, the output of the attention model is passed to the policy update of $\mathcal{M}$ to guide the model in selecting better actions. The CNN is trained using the standard backpropagation to support RL or IL decisions.


\subsection{HMM and Message Passing}
\label{Message_Passing}
As part of the probability inference view in RL, the MDP is converted to an HMM by adding the binary optimality variable $O$, where $1$ at $t$ means that $S_t$ is optimal, and $0$ otherwise. The probability of $O=1$ given the state and action taken at $t$ is inferred using a probability distribution $p$, computed using function $f$ that maps rewards to probabilities: $p(O_t = 1 | S_t, A_t) = f(r(S_t, A_t))$.

\begin{figure}[h]
    \centering
    \includegraphics[scale=0.3]{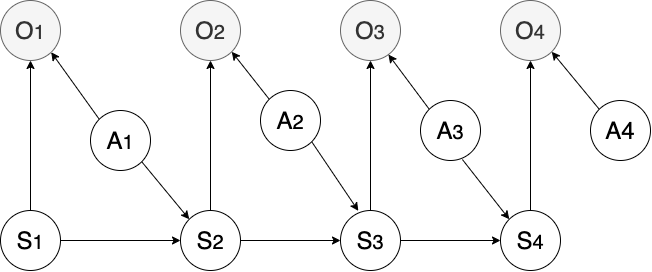}
    \caption{MDP structure as a graphical model treated as HMM}
    \label{fig:hmm}
\end{figure}

This structure is presented in Figure \ref{fig:hmm}, and is analogous to HMMs.
A backward message is computed as follows: $\beta(S_t, A_t) = p(O_{t:T} | S_t, A_t)$. In this equation, $\beta$ refers to an approximation of the cumulative reward over the trajectory from $t$ to $T$ \cite{toussaint2006probabilistic, ziebart2008maximum}. Computing the backward message is a generalization of the optimal control problem \cite{rabiner1986introduction}. A forward message is computed as follows: $\alpha(S_t, A_t) = p(O_{t_0:t-1})|S_t, A_t)p(S_t, A_t)$. This message allows the agent to examine the trajectory optimality from $0$ to $t-1$. Combining the forward and backward messages reveals information about the full trajectory, which relevant and effective for building the potential function in reward shaping.


\section{Proposed Scheme: VIN-RS}
\label{sec:proposed_scheme}
\begin{figure}[h]
    \centering
    \makebox[\textwidth][c]{\includegraphics[width=1.3\textwidth]{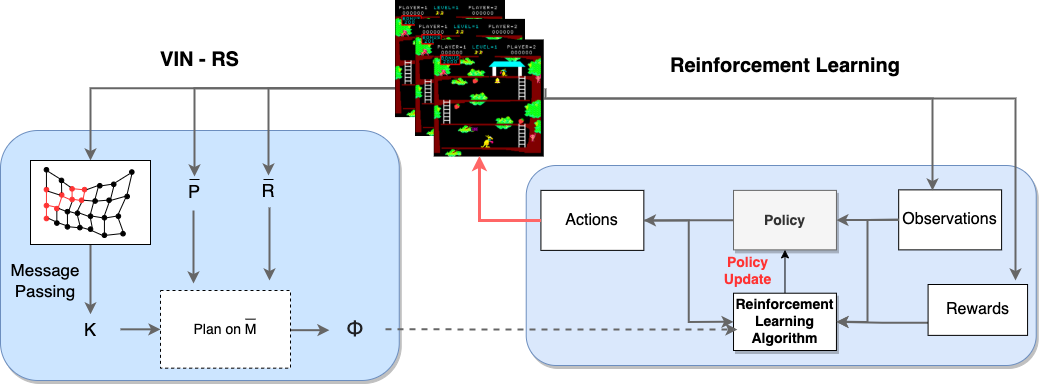}}
    \caption{An architecture incorporating the Value Iteration Network for Reward Shaping with Reinforcement Learning}
    \label{fig:vin_rs}
\end{figure}
In this section, we start first by describing the overall proposed architecture which consists of three main components: (1) VIN-RS, (2) message passing, and (3) reinforcement learning algorithm. We describe each of the components and list the steps for constructing the potential function of our reward shaping solution using CNN and focus on the advantage of incorporating the look-ahead advice mechanism. We also discuss the technique to obtain the message passing values, which are used to compute the loss for the CNN. In addition, we show an algorithm to train VIN-RS. Finally, the policy update combining the output of CNN is presented in the main reinforcement learning algorithm.

\subsection{Overall Architecture}
In Figure \ref{fig:vin_rs}, we show the overall proposed architecture that combines VIN-RS with  RL. Our solution contains two main modules, the first is VIN-RS that uses a CNN on $\bar{\mathcal{M}}$, while the second is the main RL solution for the original MDP $\mathcal{M}$.

Starting with the main component which is the RL algorithm that computes the policy $\pi$ to solve $\mathcal{M}$. The input to the RL module is the list of states which can either be images or states representation from the environment. These states are also called observations. Based on the observations, the agent selects the best action according to the policy $\pi$, then executes that action in the environment. The agent then receives the next state as well as the reward of the action taken in the environment. This information is used to update the policy using the RL algorithm. The policy then selects another action for the new observation and keeps on repeating these steps until the agent finds the optimal policy $\pi^*$. The novelty of our RL solution is that the policy update incorporates the reward shaping value, which is constructed using the output of VIN-RS.

The CNN of VIN-RS is built using three two-dimensional CNN layers. The input to CNN is a list of images captured from the environment. CNN accepts graphs, in contrast to the main RL algorithm where the input can be a state representation that combines information other than the environment images. This is similar to the case of MuJoCo environments \cite{todorov2012mujoco}. The image is passed to the first conv layer that is responsible for processing the raw image pixels. The result of the first conv layer is passed to the second one, which is responsible of producing a reward matrix $\bar{\mathcal{R}}$. This layer has two channels, one holds the old value function matrix $\bar{\mathcal{V}}$, and one holds the current rewards. In other words, states are represented as a two-dimensional grid at each timestep, and each of these states has a reward value computed using the first layer. This layer is trained and improved over time by the network through continuous weights updates. The kernel applied to this layer resembles the probability transition matrix $\bar{\mathcal{P}}$, which is also updated during the training. Applying this kernel to the first reward matrix resulted from the first conv layer will give us the state-action value function ($\bar{\mathcal{Q}}$-value). At this conv layer, there are $x$ channels, where $x$ is the number of actions in the action space $\bar{\mathcal{A}}$. Selecting the action having the maximum $\bar{\mathcal{Q}}$-value is done by applying a max-pooling for the $\bar{\mathcal{Q}}$ value of the corresponding states. The resulting matrix $\bar{\mathcal{V}}$ is passed to $\bar{\mathcal{R}}$ to be considered for the next value iteration or policy update. Furthermore, the $\bar{\mathcal{Q}}$ is flattened and a dense layer is applied to obtain the output layer. This output layer has the size of $x$, which is considered as a shaping value for each action, passed to the main policy update of the RL algorithm to update the policy using potential-based reward shaping. Producing a shaping value for each action of a given state is at the core of the look-ahead advice mechanism \cite{wiewiora2003principled}.

In order to train our CNN, we use the standard backpropagation by computing the labels using the message passing technique. As discussed in the background section, the message passing value including the forward and backward messages is considered as a signal that could accelerate learning. In order to compute the messages values, a graph of states is formed as shown in Fig. \ref{fig:cnn_vin_rs}, this graph contains only a subset of the states. Due to the fact that computing message passing is computationally expensive for large graphs, it is enough to compute this message passing for the sampled graph of states for the current training iteration of CNN. For every training episode, the graph is emptied and a new one is formed. The output of the message passing algorithm is used as the true labels for the CNN to compute the loss function. In \cite{klissarov2020reward}, message passing is implemented using a GCN; however, in this work, we overcome the limitations of using GCN, described in Section \ref{subs:related_gcn}, and apply CNN to perform value iteration and compute the message passing values. Because CNN is used to do planning in the network over $\mathcal{K}$ iterations, VIN-RS can tell if an agent state belongs to an optimal trajectory. The ability of CNN to plan using value iteration is mapped to what a message passing value represents.

\subsection{VIN-RS Module}
\begin{figure*}
    \centering
    \makebox[\textwidth][c]{\includegraphics[width=1.1\textwidth]{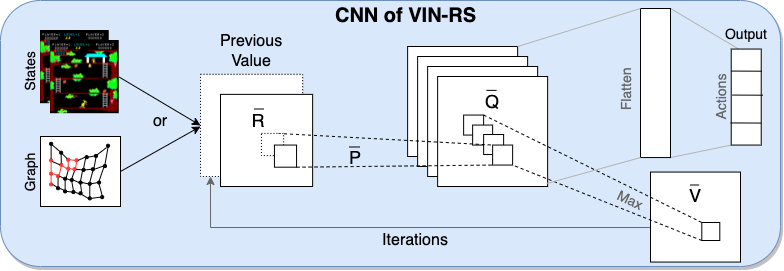}}
    \caption{The CNN architecture of the proposed VIN-RS module}
    \label{fig:cnn_vin_rs}
\end{figure*}
Our VIN-RS builds and solves $\bar{\mathcal{M}}$, where the parameters of the policy $\bar{\pi}$ include $f_r$ and $f_p$. These two functions are differentiable and learned while training the CNN. The CNN is trained using the message passing result as the label. To obtain $
\bar{\pi}^*$ for solving $\bar{\mathcal{M}}$, value iteration is applied as follows:

\begin{equation}
    \bar{\pi}^*(\bar{s}) = \text{argmax}_{\bar{a}}\bar{\mathcal{R}}(\bar{s}, \bar{a}) + \gamma \sum_{\bar{s}'} \bar{\mathcal{P}}(\bar{s}'|\bar{s}, \bar{a})\bar{V}^*(\bar{s}')
\end{equation}

\noindent where $\bar{s}'$ is the state at the next timestep. The value iteration procedure is implemented using the CNN of the proposed VIN-RS. Furthermore, we construct a VIN-RS tailored for reward shaping in the context of RL. Mainly, we train VIN-RS using the message passing values and consider its output as the shaping value that is used to update the original policy of the RL algorithm. An illustarion on the training and integration between the CNN of VIN-RS and the RL module is shown in Figure \ref{fig:vin_rs}. In addition, the CNN in VIN-RS is trained in a separate network to solve its own policy and not combined with the RL network. As shown in Figure \ref{fig:cnn_vin_rs}, the input to CNN is the list of images extracted from the environment. The input can also be a graph representation of the states from the environment, in case images are not available. Using graphs instead of images can reduce the number of states the CNN trains on; however, it makes our solution applicable to high dimensional state spaces. In other words, using grid of pixels is practical when the environment image is two-dimensional or covers the current state and end goal. Using $f_r$, which corresponds to the weights of the first conv layer, the $\bar{\mathcal{R}}$ matrix is computed. $\bar{\mathcal{R}}$ has the dimension of $l, m, n$, where $l$ is number of channels and $m$ and $n$ represent the image dimensions. The extracted reward $\bar{\mathcal{R}}$ is passed to the next conv layer, where the $\bar{\mathcal{Q}}$ values are computed. The $\bar{\mathcal{Q}}$ conv layer contains $x$ channels, or a channel for each action in $\bar{\mathcal{M}}$. $\bar{\mathcal{Q}}_{\bar{a}, i', j'}$ represents the $\bar{\mathcal{Q}}$-value for a state defined at positions between $i, i'$ and $j, j'$ respectively for a particular action $\bar{a}$. $\bar{\mathcal{Q}}_{\bar{a}, i', j'}$ is computed as follows:
\begin{equation}
    \bar{\mathcal{Q}}_{\bar{a}, i', j'} = \sum_{l, i, j} W_{l, i, j}^{\bar{a}}\bar{\mathcal{R}}_{l, i'-i, j'-j}
    \label{eq:q_value}
\end{equation}
In Equation \ref{eq:q_value}, the reward matrix is multiplied by the weights or a representation of the transition matrix $\bar{\mathcal{P}}$. From the resulting $\bar{\mathcal{Q}}$ matrix, we apply max-pooling by selecting the highest $\bar{\mathcal{Q}}$-value from the list of actions or channels to obtain $\bar{\mathcal{V}}$. An element in $\bar{\mathcal{V}}$ at $i,j$ is:
\begin{equation}
    \bar{\mathcal{V}}_{i,j} = \text{max}_{\bar{a}}\bar{\mathcal{Q}}(\bar{a}, i, j)
\end{equation}
\noindent where $\bar{\mathcal{Q}}(\bar{a}, i, j)$ is the result of the $\bar{\mathcal{Q}}$ function for a given state and action. Following the computation of $\bar{\mathcal{V}}$, we update the first channel of the $\bar{\mathcal{R}}$ matrix. In addition, a dense layer is applied after flattening $\bar{\mathcal{Q}}$. Finally, a fully connected output layer is added, which results in the shaping values for each action. In order to obtain the true labels for the shaping value, we apply the message passing mechanism on the extracted graph from the observations, and pass this value to the loss of CNN. A forward pass in this CNN is considered as a single value iteration. Assuming that the number of steps required by the agent to reach the goal from the current state is $\mathcal{K}$, then the ideal number of value iterations required in CNN for value function updates is $\mathcal{K}$. After each iteration, $\bar{\mathcal{V}}$ is calculated using $\bar{\mathcal{Q}}$ and appended to $\bar{\mathcal{R}}$ for the next iteration, as shown in Figure \ref{fig:cnn_vin_rs}. In Algorithm \ref{alg:cnn_vin_rs}, we show a pseudo-code of the training steps in our proposed CNN.

\begin{algorithm}[!htbp]
    \caption{A training step with $\mathcal{K}$ value iterations of CNN in VIN-RS}
    \label{alg:cnn_vin_rs}
    \begin{flushleft}
    \textbf{Input}: $X$: a list of images sampled from the environment.\\ $\mathcal{G}$: a graph constructured from the sampled images or the list of states encountered.\\
    $cnnH$: Conv layer that processes the image pixels\\
    $cnnR$: Conv layer that computes the $\bar{\mathcal{R}}$ matrix\\
    $cnnQ$: Conv layer that computes the $\bar{\mathcal{Q}}$ matrix\\
    $wV$: Initialize weights for computing $\bar{\mathcal{Q}}$ matrix\\
    $Fn$: Fully connected layer\\
    $Opt$: Output layer\\
    \textbf{Output}: $\phi$: shaping value for the corresponding images/states.
    \end{flushleft}
    \begin{algorithmic}[1]
         \State $p = \text{Normalize}(X)$ \Comment{or $G$ as input when applicable}
         \State $h = cnnH(p)$
         \State $r = cnnR(h)$
         \State $q = cnnQ(r)$
         \State $v = \text{Max}(q)$ \Comment{Get the maximum or apply max pooling}
         \For{$i = 1, \dots \mathcal{K}$} \Comment{Perform $\mathcal{K}$ value iterations}
            \State $q = \text{EVALUATE\_Q}(r, v)$
            \State $v = \text{Max}(q)$
         \EndFor
         \State $q = \text{EVALUATE\_Q}(r, v)$
         \State $v = \text{Max}(q)$
         \State $flatten\_q = \text{Reshape}(q)$ \Comment{Flatten the matrix v}
         \State $fn = Fn(flatten\_q)$
         \State $\phi = Opt(fn)$
         \State $label = \text{Message\_Passing}(\mathcal{G})$
         \State $loss = \mathcal{L}(label, \phi)$
         \State $backpropagate()$
         \State $update\_gradient()$
         \Procedure{\text{evaluate\_q}}{$(r, v, cnnQ, wV)$} \Comment{Perform value iteration}
            \State $rv = \text{Concat}(r, v)$ \Comment{Concatenate $r$ and $v$}
            \State $wQwV = \text{Concat}(cnnQ\text{.weights}, wV)$
            \State $q = Conv(rv, wQwV)$ \Comment{Apply a Convolutional layer}
            \State $\text{return } q$
        \EndProcedure
    \end{algorithmic}
\end{algorithm}
The pseudo-code of Algorithm \ref{alg:cnn_vin_rs} presents how the CNN of VIN-RS is trained for a single step. Lines 1-5 perform a forward pass in the algorithm. Lines 6-10 perform $\mathcal{K}$ value iterations. Lines 11-13 flatten the $\bar{\mathcal{Q}}$ matrix and obtain the output $\phi$. Lines 14-15 compute the message passing values and compute the loss. Line 16 performs backpropagation to compute the gradients of the network weights. Line 17 updates the weights based on the computed gradient. 

\subsection{Loss Function: Message passing}
Following the success of using message passing for reward shaping, we propose training the CNN of VIN-RS using the message passing values as the true labels. Using the probability inference view of RL, a solution is to find the distribution of the optimality variable $\mathcal{O}=1$ for a given state and action. This structure is analogous to HMM, where forward and backward messages can be used to compute this probability distribution. Combining the forward and backward messages results in a policy that looks backward and forward in time. In other words, the resulting values from message passing tells if a state belongs to an optimal trajectory. More details about the messages calculation is presented in Section \ref{Message_Passing}. Thus, the combined forward and backward messages ($\alpha$ and $\beta$) are expressed as follows:
\begin{equation}
    \label{eq:message_passing}
    p(O_t | {S}_t, A_t) \simeq \alpha(S_t, A_t)\beta(S_t, A_t)
\end{equation}
Thus, the potential function is expressed as $\phi_{\alpha, \beta} = \alpha(S_t, A_t)\times\beta(S_t, A_t)$. Compared to VIN, a CNN can also produce optimal plans/trajectories for the agent on $\bar{\mathcal{M}}$. Based on this observation, we propose VIN-RS that incorporates the message passing results in the training process to produce the shaping values.

In VIN-RS, we benefit from message passing to compute the loss of CNN. To compute those messages, base and recursive losses are required. In our loss formulation, the state and action are passed as input when computing the loss. Such formulation boosts the performance of VIN-RS by applying the look-ahead advice naturally in the CNN implementation, due to the fact that each channel in $\bar{\mathcal{Q}}$ is a Q-function for each action in the environment. Therefore, the loss function is computed in two steps as follows:
\begin{equation}
    \label{loss_equation}
    \mathcal{L} = \mathcal{L}_0(\bar{\mathbf{S}}, \bar{\mathbf{A}}) + \eta \mathcal{L}_{rec}(\bar{\mathcal{S}}, \bar{\mathcal{A}})
\end{equation}
where $\mathcal{L}_0$ is the base case, $\mathcal{L}_{rec}$ is the recursive case, $\bar{\mathbf{S}}$ and $\bar{\mathbf{A}}$ are the lists of base case states and actions respectively in $\bar{\mathcal{M}}$, and $\bar{\mathcal{S}}$ and $\bar{\mathcal{A}}$ are the states and actions retrieved from the graph of experiences. Noting that the base states are composed of the rewarding states. In VIN-RS, we consider states that are used to form the graph $\mathcal{G}$, where each state in this graph is an images of the environment. Furthermore, the base loss is computed using the rewarding states only in $\mathcal{G}$, in order to extract information only from important states. The base loss is the usual supervised loss that considers the true and predicted labels. 

\begin{equation}
\begin{aligned}
    &\mathcal{L}_0 = H(p(\bar{O} | \bar{\mathbf{S}}, \bar{\mathbf{A}}), \phi(\bar{\mathbf{S}}, \bar{\mathbf{A}})) = \\
    &\sum_{s, a\in \bar{\mathbf{S}}, \bar{\mathbf{A}}}p(\bar{O}|s, a)\text{log}(\phi(s, a))
    \label{eq:loss_base}
    \end{aligned}
\end{equation}
The recursive loss written as $\mathcal{L}_{rec}$ is computed by aggregating the messages with the neighboring states using the adjacency matrix of the graph $\mathcal{G}$. $\mathcal{L}_{rec}$ is formulated as follows:
\begin{equation}
    \mathcal{L}_{rec} = \sum_{i=1}^{||d||}\sum_{j=1}^{||e||} \mathfrak{A}_{i, j} ||\phi(\bar{\mathcal{S}}_i, \bar{\mathcal{A}}_i) - \phi(\bar{\mathcal{S}}_j, \bar{\mathcal{A}}_j)||^2
\label{eq:loss_rec}
\end{equation}
In Equation \ref{eq:loss_rec}, $d$ and $e$ are the list of states and corresponding neighbors respectively. In addition, $\mathfrak{A}$ is the adjacency matrix. Getting $\phi$ for a given state and action is at the output layer $Opt$ of CNN for VIN-RS. Compared to \cite{klissarov2020reward}, our loss function considers both the states and actions for activating the look-ahead advice mechanism.

\subsection{Look-Ahead Advice}
\label{look_ahead}
The Look-ahead advice mechanism proposed in \cite{wiewiora2003principled} suggests considering the action as part of the potential-based reward shaping function. Advising on specific actions is a more rigorous method taken at the level of actions instead of being general for the whole state. The shaping function produced by CNN in VIN-RS after applying the look-ahead advice takes the following form:
\begin{equation}
    F(\bar{s}, \bar{a}, \bar{s}', \bar{a}') = \gamma\phi(\bar{s}', \bar{a}') - \phi(\bar{s}, \bar{a})
    \label{req:look_ahead_potential}
\end{equation}
where $\phi$ is the potential shaping function that considers states from $\mathcal{S}$ and actions from $\mathcal{A}$ to result in a scalar value. Hence, the updated shaping function considering the action taken becomes:
\begin{equation}
    R(s, a, s', a') = r(s, a) + F(s, a, s', a')
    \label{req:look_ahead_reward_shaping}
\end{equation}
By augmenting the action values, the shaping function could potentially speed the learning speed further. Therefore, we propose adding the look-ahead advice in the design of VIN-RS, which is naturally embedded at the $\bar{\mathcal{Q}}$ layer.

\subsection{Training RL with VIN-RS}
VIN-RS is able to produce shaping values that can accelerate learning and overcome the limitation of existing solutions by processing full images of the environment or extracted graphs of states. After training the CNN of VIN-RS, the resulting shaping values for each state and action are passed to the RL algorithm for training and policy update. In Algorithm \ref{alg:rl_vin_rs}, we show the steps followed to train RL and benefit from the shaping value to obtain the $\mathcal{Q}_{comb}$, which is a combination of the original $\mathcal{Q}$ value and the one obtained using Equation \ref{eq:reward_shaping} with $\phi$ from CNN as the shaping value. The algorithm starts by initializing the CNN and RL networks, as well as an empty graph $\mathcal{G}$ to hold the list of transitions. In each epoch of training, for a number of iterations $T$ (for each trajectory followed by the agent), images are stored to later train the CNN. In addition, the list of transitions are stored in graph $\mathcal{G}$ to later compute the loss of CNN. For every $N$ episodes, the CNN is trained with the sampled images and graphs of transitions. Training the CNN every $N$ episodes is more efficient and reduces the runtime of using VIN-RS in combination with the RL solution to speed learning. The loss is computed using Equation \ref{loss_equation} and the CNN is trained for $\mathcal{K}$ iterations following the steps of Algorithm \ref{alg:cnn_vin_rs}. Noting that if images are not available for training the CNN, the graph $\mathcal{G}$ can be used instead.

The combined value function with reward shaping is expressed as:
\begin{equation}
    \begin{aligned}
    &\mathcal{Q}_{comb}^\pi (s, a)= \alpha \mathcal{Q}^\pi (s, a)+ (1 - \alpha) \bar{\mathcal{Q}}^{\bar{\pi}}_{\phi}(s, a)\\
    &\text{where } \bar{\mathcal{Q}}^{\bar{\pi}}_{\phi}(s, a)= \mathbf{E}_{(s, a)}[\sum_t \gamma^tr(S_t, A_t) + \gamma\phi(S_{t+1},\\
    &A_{t+1}) - \phi(S_{t}, A_{t})]
\end{aligned}
\end{equation}

$\phi(S_{t}, A_{t})$ is the shaping value at the output layer $Opt$ of the CNN for state $S_t$ and action $A_t$. Moreover, $\alpha$ is a hyperparameter the amount of the reward shaping value considered for updating the state-action value function. At the end of an epoch, the graph $\mathcal{G}$ can be emptied. 

\begin{algorithm}[!htbp]
    \caption{Training RL with VIN-RS}
    \begin{algorithmic}[1]
        \State Create the CNN network for VIN-RS
        \State Create empty graph $\mathcal{G}$
        \State Create the RL networks
        \For{Episode=0,1,2, \dots}
            \For{$t = 1, 2, \dots, T$}
                \State Store images of all transitions
                \State Perform the best action based on $\pi$
                \State Get the state and reward from the environment
                \State Add the transition to graph $\mathcal{G}$
            \EndFor
            \If{mod(Episode, $N$)}
                \State Pass the list of images to CNN
                \State Compute the loss for CNN
                using Equation \ref{loss_equation}
                \State Train CNN following Algorithm \ref{alg:cnn_vin_rs} for $\mathcal{K}$ iterations
            \EndIf
            \State Obtain $\phi$ for the list of states and actions
            \State $\mathcal{Q}_{comb}^\pi = \alpha \mathcal{Q}^\pi + (1 - \alpha) \bar{\mathcal{Q}}^{\bar{\pi}}_{\phi}$
            \State Train RL networks by updating the policy to maximize $E_\pi [\nabla log \pi(A_t | S_t) \mathcal{Q}_{comb}^\pi(S_t, A_t)]$
            \State Reset $\mathcal{G}$ to empty graph (optional)
        \EndFor
    \end{algorithmic}
    \label{alg:rl_vin_rs}
\end{algorithm}

\section{Experiments}
In this section, the evaluation consists of experiments on two environments with discrete and continuous state and control. We use twenty Atari 2600 games from the Gym environment and four games from MuJoCo. In order to analyse the performance of VIN-RS and illustrate its advantage, we compare with the Proximal Policy Optimization (PPO) \cite{schulman2017proximal}; using GCN (denoted as $\phi_{GCN}$) as the shaping function with the graph Laplacian as the filter \cite{klissarov2020reward}; LIRPG \cite{zheng2018learning}; RND \cite{burda2018exploration}; and ICM \cite{pathak2017curiosity}. Details about the implementation and machines used are provided in the next subsection. First, we analyze the time complexity of VIN-RS compared to PPO and GCN. Afterward, we evaluate the performance of VIN-RS in different games for the Atari 2600 \cite{brockman2016openai}, and MuJoCo \cite{todorov2012mujoco} environments compared to various baselines.
\label{sec:experiments}


\subsection{Implementation and Setup}
The source code is written in the Python programming language. In our implementation, we used the PyTorch library to build our VIN-RS and combine it with the implementation of the Actor Critic (A2C) and Proximal Policy Optimization (PPO) algorithm. We also utilize the OpanAi Gym \cite{brockman2016openai} and MuJoCo \cite{todorov2012mujoco} libraries to simulate the environments of all the games. Images are passed to CNN to train in both the Atari and MuJoCo environments. The state representation of the MuJoCo games contains additional information about the state and not only the raw pixels. Therefore, we utilized the camera option in the MuJoCo package to build the CNN input. In case there is no option to get images from the environment, a graph of states can be used as input to CNN. The same graph of states is used to compute the loss function, through message passing, when training the CNN. Passing images as input results in more information about the environment, which further improves the performance of CNN. As described in Section \ref{sec:proposed_scheme}, the look-ahead advice mechanism is naturally embedded within the VIN-RS design and implementation, which offers a level of advantage for the proposed scheme compared to existing baselines. Our source code is available on GitHub and will be shared after publication.

For each run, a single GPU (NVIDIA V100 Volta (16GB HBM2 memory)) and eight CPUs (Intel E5-2683 v4 Broadwell @ 2.1GHz) were used with 32 GB of RAM.
Details about the network configuration of each environment are provided in Sections \ref{sec:atari} and \ref{sec:mujoco}.

\subsection{Complexity}

In this section, we show the results of our analysis in terms of runtime of the proposed VIN-RS when combined with PPO compared to GCN and vanilla PPO. In Table \ref{tab:fps}, we show the number of frames processed per second (FPS) using each of the solutions. As presented in Algorithm \ref{alg:cnn_vin_rs}, the CNN of VIN-RS is only trained every $N$ episodes, that's why the FPS is very close compared to the other solutions. Therefore, when comparing the performance of VIN-RS to the other baselines, it is enough to compare the cumulative steps to converge or the average reward achieved over the number of iterations.
\begin{wraptable}{r}{3.2cm}
    \begin{tabular}{c|c}
        Method & FPS\\
        \hline
        \hline
        PPO & 1126\\
        GCN & 1054\\
        VIN-RS & 1051
    \end{tabular}
    \caption{Frame Per Second (FPS) evaluated on Atari 2600}
    \label{tab:fps}
\end{wraptable}
Training VIN-RS consumes additional time every couple of episodes due to the added computation of the loss using Equation \ref{loss_equation}, in addition to the steps of training CNN following Algorithm \ref{alg:cnn_vin_rs}. Therefore, VIN-RS has comparable execution time compared to PPO and GCN, thus not affecting the speed of learning. Hence, studying the number of iterations to convergence has the same effect as measuring the time in seconds when comparing to different baselines in the following subsections.

\subsection{Performance Analysis}
In our evaluations, we consider three different environments to test the performance of the proposed VIN-RS. First is the Tabular with the four rooms game, second is the Atari 2600, and third is the MuJoCo. All these environments are similar to the evaluation criteria followed in \cite{klissarov2020reward} that proposes the use of GCN to perform message passing and predict the shaping value.
Even-though the $\phi_{CNN}$ (using VIN-RS) approach for reward shaping achieves considerable improvement over the PPO algorithm in \cite{klissarov2020reward}, we still provide a comparison with A2C and PPO as baselines.

\subsubsection{\textbf{Tabular Learning}}
\begin{figure}[!htb]
    \centering
    \begin{subfigure}{0.25\textwidth}
        \includegraphics[width=\textwidth]{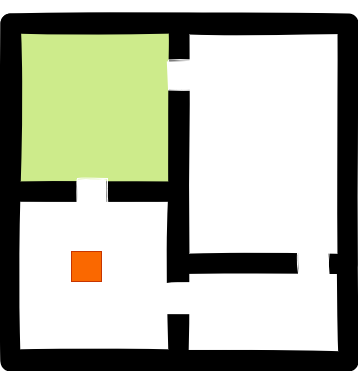}
        \caption{Four Rooms}
        \label{fig:fr}
    \end{subfigure}
    \begin{subfigure}{0.6\textwidth}
        \includegraphics[width=\textwidth]{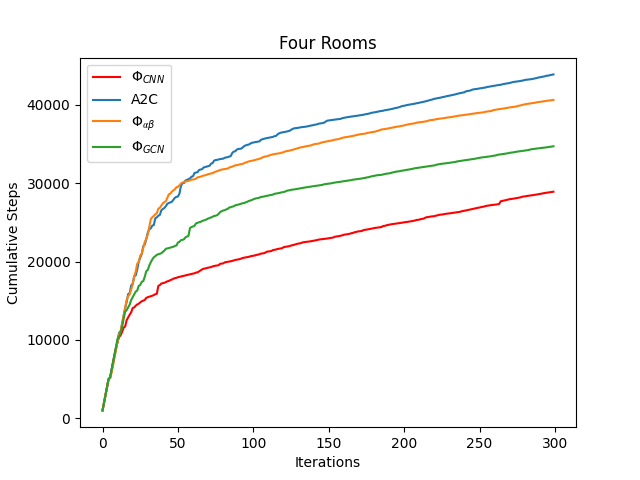}
        \caption{Convergence Speed}
        \label{fig:fr_res}
    \end{subfigure}
    \caption{Cumulative steps over the number of iterations in Four Rooms}
    \label{fig:fr_1}
\end{figure}

\begin{figure}[!htb]
    \centering
    \begin{subfigure}{0.25\textwidth}
        \includegraphics[width=\textwidth]{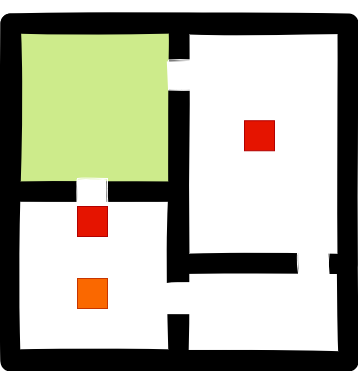}
        \caption{Four Rooms Traps}
        \label{fig:fr2}
    \end{subfigure}
    \begin{subfigure}{0.6\textwidth}
        \includegraphics[width=\textwidth]{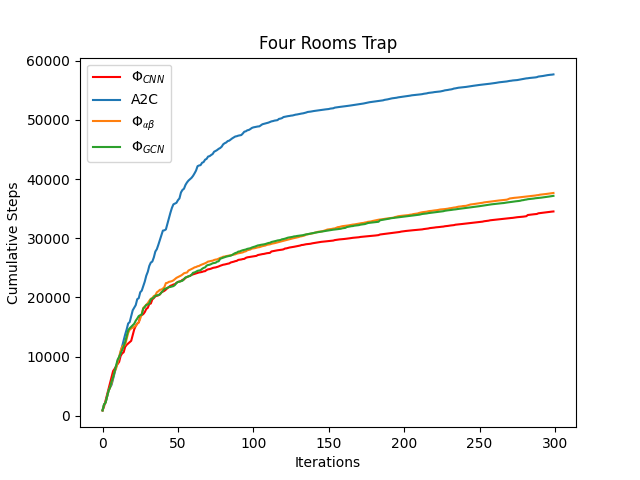}
        \caption{Convergence speed}
        \label{fig:fr2_res}
    \end{subfigure}
    \caption{Cumulative steps over the number of iterations in Four Rooms Traps}
    \label{fig:fr_2}
\end{figure}

In this experiment, we present two setups of the Four Rooms game to evaluate the performance of VIN-RS $\phi_{CNN}$ that uses A2C. The analysis conducted on VIN-RS is compared to A2C, $\phi_{GCN}$, and $\phi_{\alpha\beta}$. In $\phi_{\alpha\beta}$, message passing is computed due to the small environment spaces. Furthermore, $\lambda$-return is used as the critic part of A2C. A tabular RL solution is enough for the four rooms game shown in Figure \ref{fig:fr}. In such an environment, it is possible to compute the actual message passing value $\phi_{\alpha\beta}$. However, in larger environment sizes and dimensions, it is not feasible to compute those messages. The two games we evaluate are the Four Rooms and its variant Four Rooms Traps, where negative rewards are scattered across the four rooms as traps. Moreover, the exploration rate is maintained by setting the probability of random action selection to 0.1. The results showing the cumulative steps are presented in Figures \ref{fig:fr_1} and \ref{fig:fr_2}.
As shown in Figure \ref{fig:fr_1}, $\phi_{CNN}$ has faster convergence speed indicated by less number of cumulative steps compared to A2C, $\phi_{\alpha\beta}$, and $\phi_{GCN}$. In particular, after 300 episodes, $\phi_{CNN}$ In this particular game, VIN-RS helps the agent plan the optimum trajectory by selecting the shortest path to reach the goal with a smaller number of training iterations. The planning factor is not possible when using message passing through GCN as the potential function, because (1) a subset of the states is only considered when training, and (2) value iteration is not performed to ensure planning.\\

\subsubsection{\textbf{Atari 2600}}
\label{sec:atari}
The Gym library offers environments for twenty different Atari 2600 games. The main property of these games is that the action space is discrete. In this section, we evaluate the performance of our proposed solution $\phi_{CNN}$ in each of the games compared to four baselines, which are PPO, $\phi_{GCN}$, LIRPG, RND, and ICM. 
In terms of VIN-RS implementation, we use the states which are raw pixel representation as input to the CNN. Furthermore, the number of channels at the $\bar{\mathcal{Q}}$ layer is equal to the number of actions of the game. 

We experiment with the twenty different Atari 2600 games using $\phi_{CNN}$, $\phi_{GCN}$, LIRPG, RND, ICM, and PPO. We execute each algorithm on every game for ten million steps, the same parameters are used as shown in Table \ref{tab:atari_param}.

\begin{table}[H]
    \centering
    \caption{VIN-RS and RL configuration for the Atari 2600 games}
    \begin{tabular}{c|c}
        Hyperparameter & Value \\
        \hline
        Learning rate & 2.5e-4\\
        $\gamma$ & 0.99\\
        $\lambda$ & 0.95\\
        Entropy Coefficient & 0.01\\
        PPO steps & 128\\
        PPO Clipping Value & 0.1\\
        \# of minibatches & 4\\
        \# of processes & 8\\
        CNN: $\alpha$ & 0.9\\
        CNN: $\eta$ & 1e1\\
        GCN: $\alpha$ & 0.9\\
        GCN: $\eta$ & 1e1\\
    \end{tabular}
    \label{tab:atari_param}
\end{table}

\begin{figure*}[!htbp]
    \centering
    \begin{subfigure}{0.49\textwidth}
        \caption{$\phi_{CNN}$}
        \includegraphics[width=\textwidth]{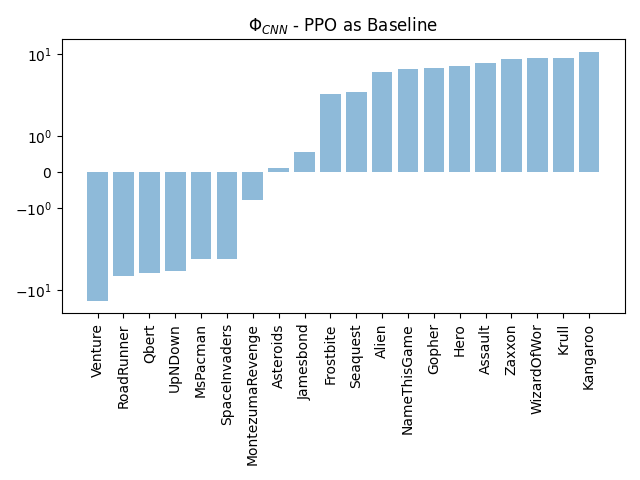}
        \label{fig:res_ppo_atari}
    \end{subfigure}
    \hfill
    \begin{subfigure}{0.49\textwidth}
        \caption{$\phi_{GCN}$}
        \includegraphics[width=\textwidth]{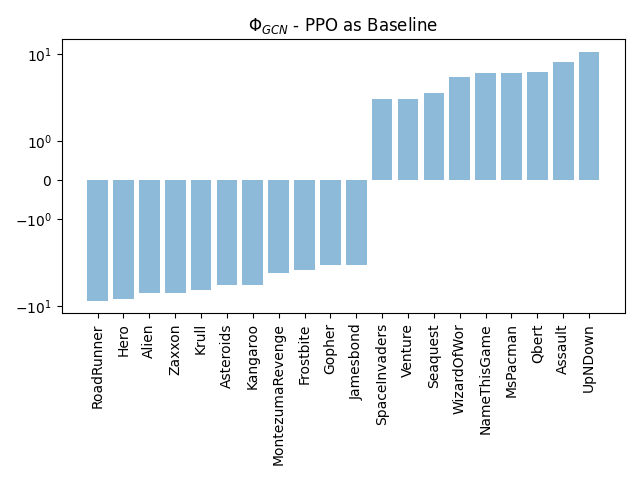}
        \label{fig:phi_gcn}
    \end{subfigure}
    \hfill
    \begin{subfigure}{0.49\textwidth}
        \caption{LIRPG}
        \includegraphics[width=\linewidth]{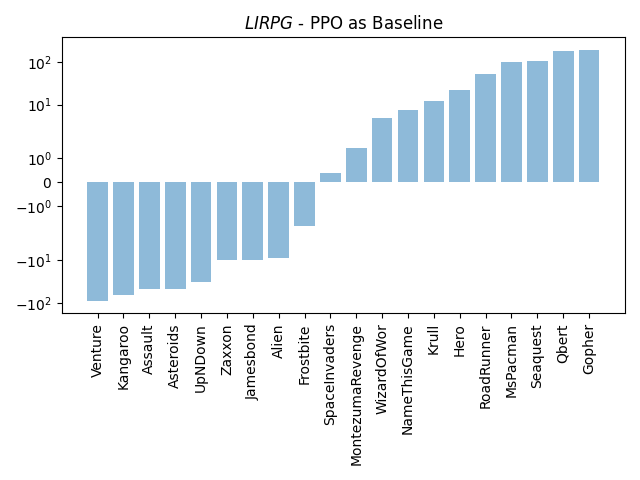}
        \label{fig:LIRPG}
    \end{subfigure}
    \hfill
    \begin{subfigure}{0.49\textwidth}
        \caption{RND}
        \includegraphics[width=\linewidth]{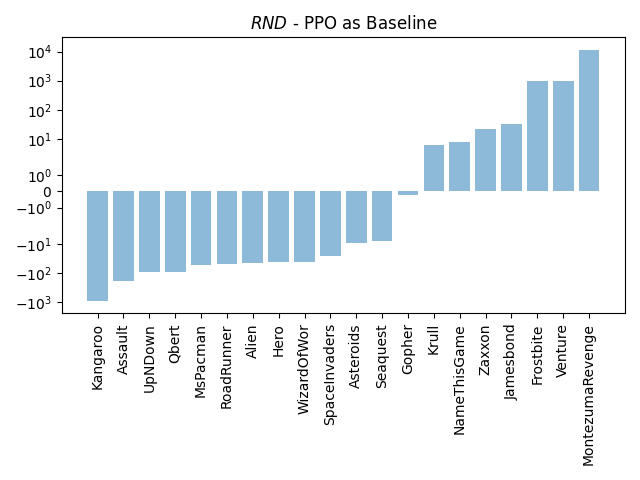}
        \label{fig:rnd}
    \end{subfigure}
    \hfill
    \begin{subfigure}{0.49\textwidth}
        \caption{ICM}
        \includegraphics[width=\linewidth]{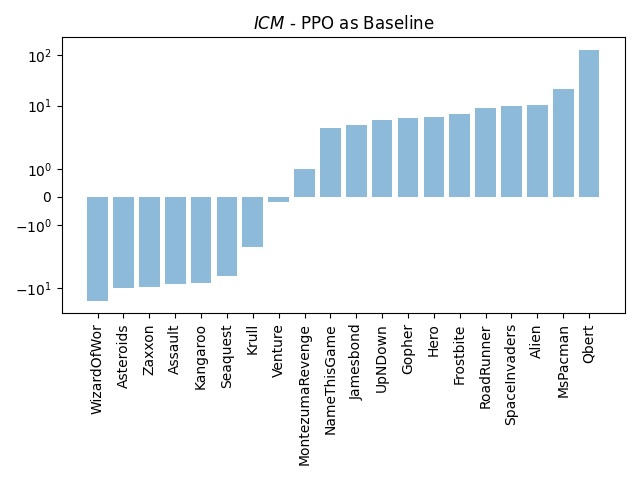}
        \label{fig:ICM}
    \end{subfigure}
    \caption{Performance comparison of each learning and reward shaping algorithm on Atari games in log scale over ppo}
    \label{fig:res_atari}
\end{figure*}

In Figure \ref{fig:res_ppo_atari}, we show the improvement achieved by $\phi_{CNN}$ using VIN-RS over PPO. In addition, we present the improvement of each of the baselines $\phi_{GCN}$, LIRPG, RND, and ICM compared to PPO in Figures \ref{fig:phi_gcn}, \ref{fig:LIRPG}, \ref{fig:rnd}, and \ref{fig:ICM} respectively. The results in these figures are shown in logarithmic scale. The mean difference between each solution and PPO is computed, then a log is applied.

As shown in Figure \ref{fig:res_ppo_atari}, the proposed $\phi_{CNN}$ is able to improve the performance of 14 different Atari games compared to PPO as a baseline. In the games of Venture, Qbert, UpNDown, MsPacman, and SpaceInvaders, $\phi_{CNN}$ is not performing well compared to PPO. In contrast, the performance of $\phi_{GCN}$ in this game is better compared to PPO. On the other hand, in most of the other games, $\phi_{CNN}$ is outperforming $\phi_{GCN}$ and improving over PPO. These information are extracted from Figures \ref{fig:res_ppo_atari} and \ref{fig:phi_gcn}. Furthermore, the proposed $\phi_{CNN}$ has the best results in terms of the number of games where the performance is better than PPO compared to the rest of the baselines. More specifically, the number of games with an improvement of PPO are classified as $\phi_{CNN}$: 13, $\phi_{GCN}: 9$, LIRPG: 11, RND: 7, ICM: 12.\\

\begin{figure*}[!htbp]
    \centering
    \begin{subfigure}{0.242\textwidth}
        \caption{Alien}
        \includegraphics[width=\textwidth]{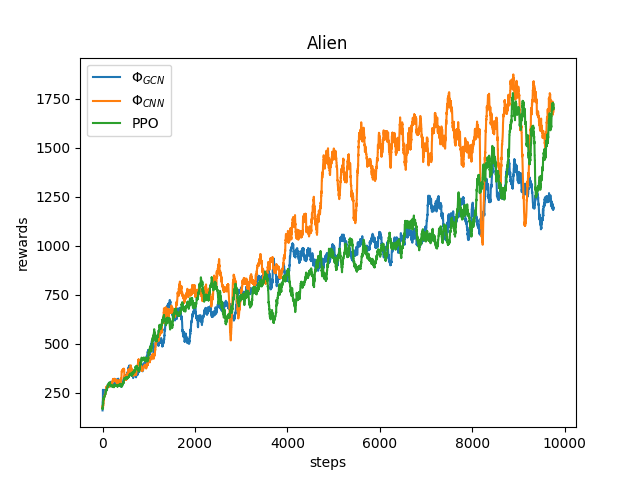}
    \end{subfigure}
    \begin{subfigure}{0.242\textwidth}
        \caption{Assault}
        \includegraphics[width=\textwidth]{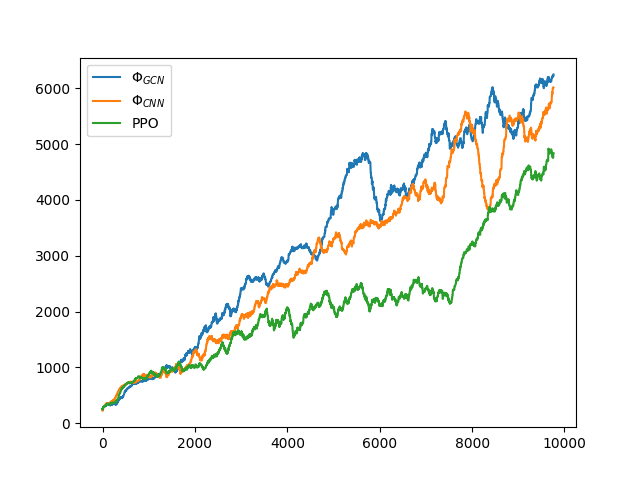}
    \end{subfigure}
    \begin{subfigure}{0.242\textwidth}
        \caption{Asteroids}
        \includegraphics[width=\linewidth]{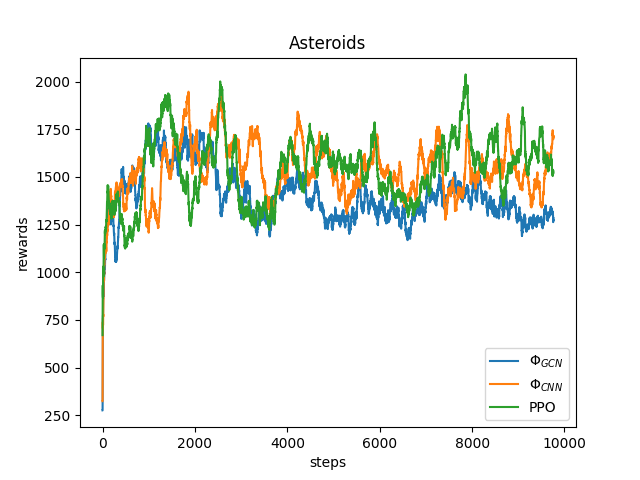}
    \end{subfigure}
    \begin{subfigure}{0.242\textwidth}
        \caption{Frosbite}
        \includegraphics[width=\linewidth]{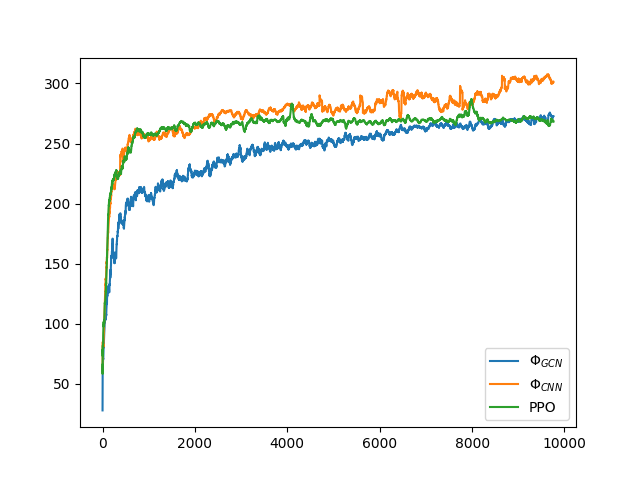}
    \end{subfigure}
    
    \begin{subfigure}{0.242\textwidth}
        \caption{Gopher}
        \includegraphics[width=\textwidth]{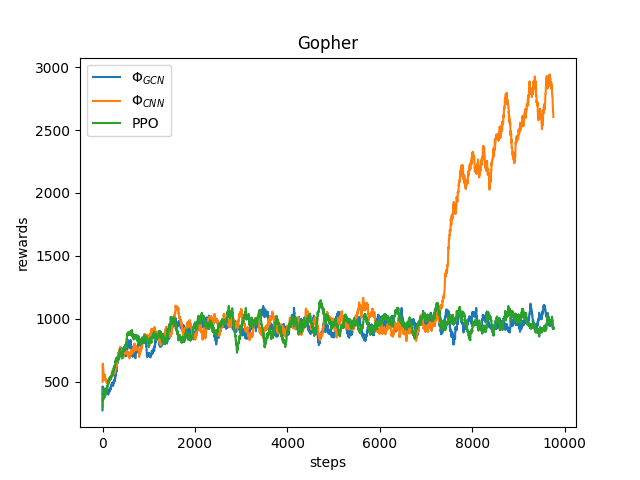}
    \end{subfigure}
    \begin{subfigure}{0.242\textwidth}
        \caption{Jamesbond}
        \includegraphics[width=\textwidth]{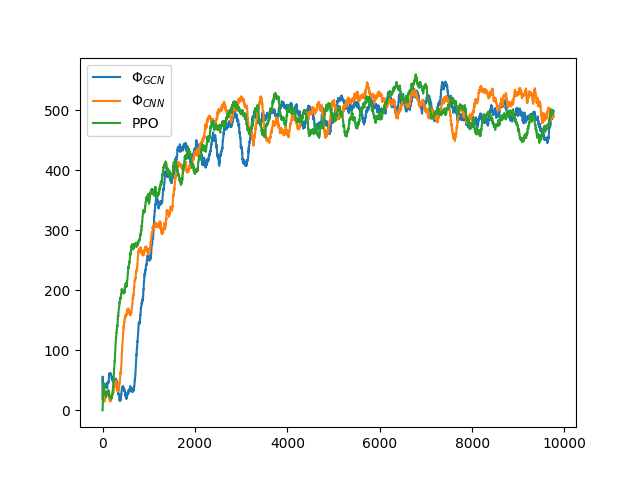}
    \end{subfigure}
    \begin{subfigure}{0.242\textwidth}
        \caption{Hero}
        \includegraphics[width=\linewidth]{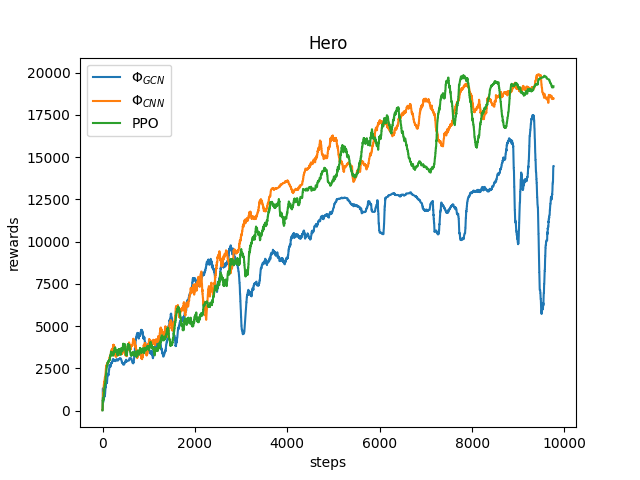}
    \end{subfigure}
    \begin{subfigure}{0.242\textwidth}
        \caption{Kangaroo}
        \includegraphics[width=\linewidth]{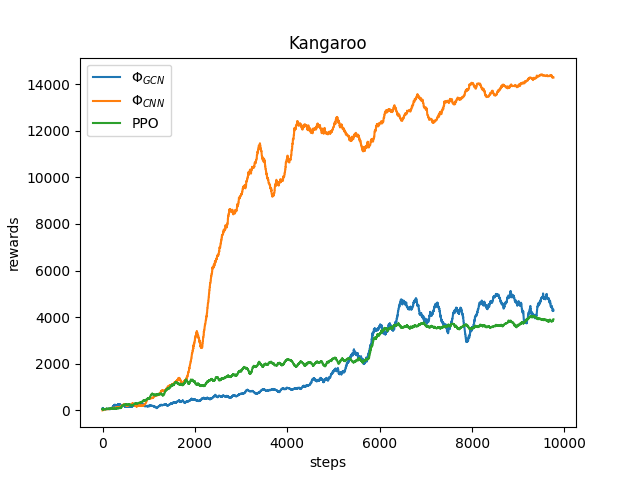}
    \end{subfigure}
    
    \begin{subfigure}{0.242\textwidth}
        \caption{Krull}
        \includegraphics[width=\textwidth]{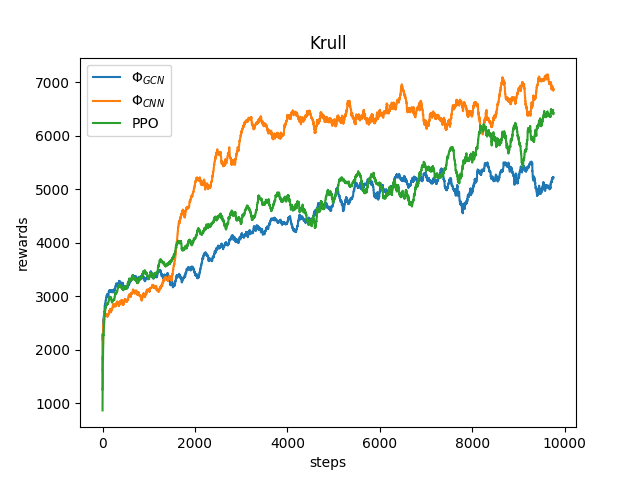}
    \end{subfigure}
    \begin{subfigure}{0.242\textwidth}
        \caption{MontezumaRevenge}
        \includegraphics[width=\textwidth]{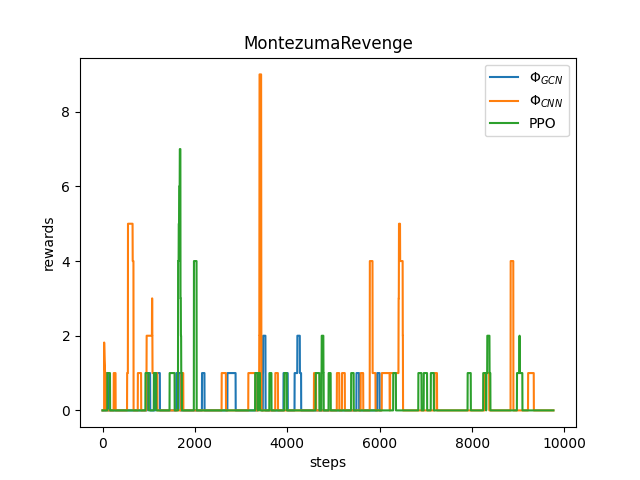}
    \end{subfigure}
    \begin{subfigure}{0.242\textwidth}
        \caption{MsPacman}
        \includegraphics[width=\linewidth]{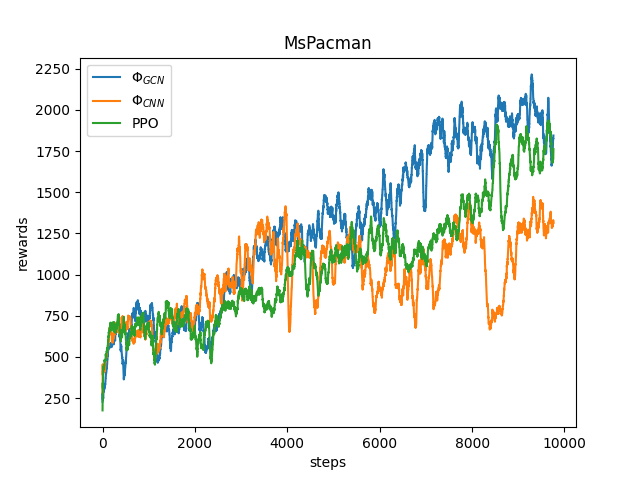}
    \end{subfigure}
    \begin{subfigure}{0.242\textwidth}
        \caption{NameThisGame}
        \includegraphics[width=\linewidth]{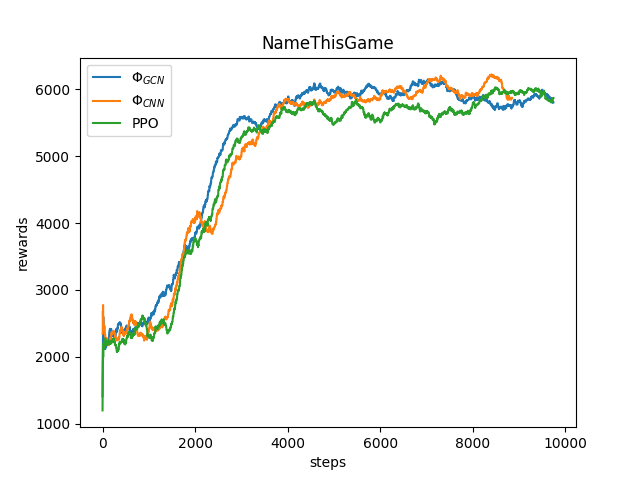}
    \end{subfigure}
    
    \begin{subfigure}{0.242\textwidth}
        \caption{Qbert}
        \includegraphics[width=\textwidth]{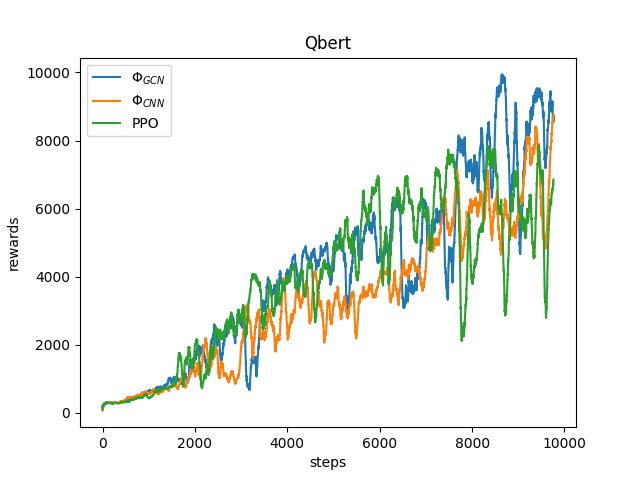}
    \end{subfigure}
    \begin{subfigure}{0.242\textwidth}
        \caption{RoadRunner}
        \includegraphics[width=\textwidth]{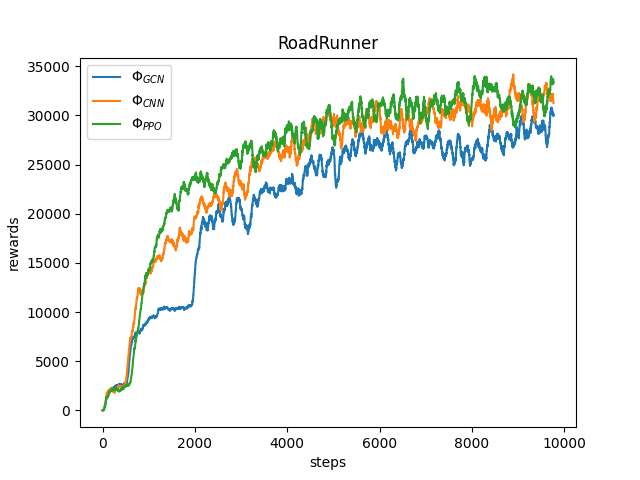}
    \end{subfigure}
    \begin{subfigure}{0.242\textwidth}
        \caption{Seaquest}
        \includegraphics[width=\linewidth]{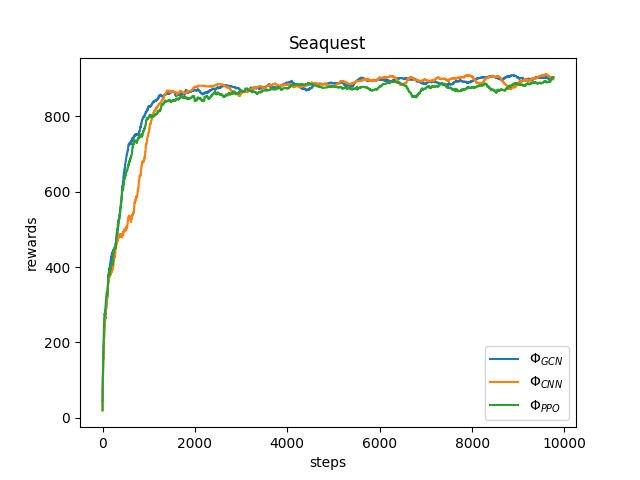}
    \end{subfigure}
    \begin{subfigure}{0.242\textwidth}
        \caption{SpaceInvaders}
        \includegraphics[width=\linewidth]{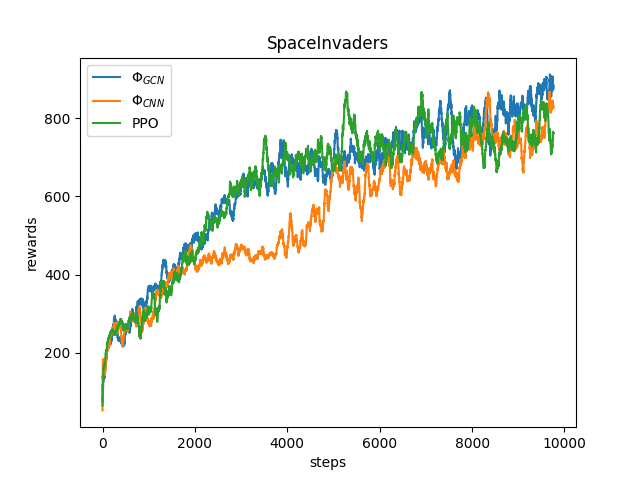}
    \end{subfigure}
    
    \begin{subfigure}{0.242\textwidth}
        \caption{UpNDown}
        \includegraphics[width=\textwidth]{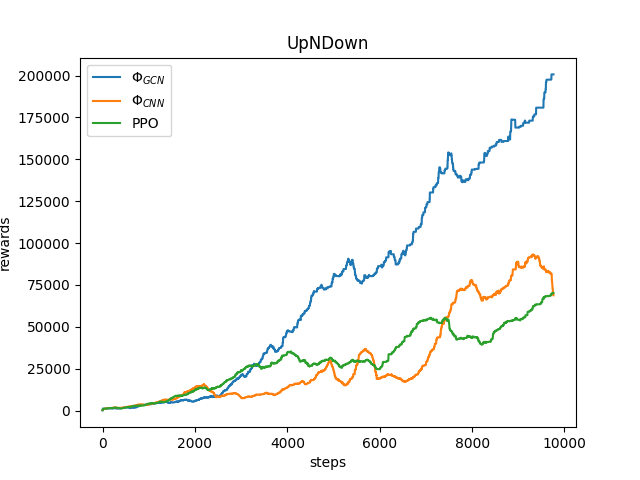}
    \end{subfigure}
    \begin{subfigure}{0.242\textwidth}
        \caption{Venture}
        \includegraphics[width=\textwidth]{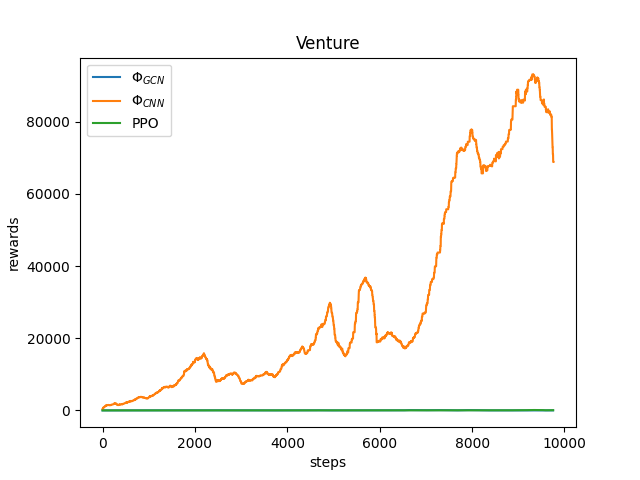}
    \end{subfigure}
    \begin{subfigure}{0.242\textwidth}
        \caption{WizardOfWar}
        \includegraphics[width=\linewidth]{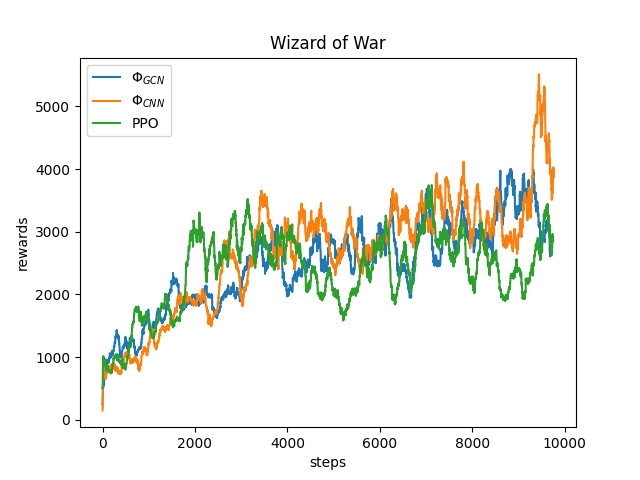}
    \end{subfigure}
    \begin{subfigure}{0.242\textwidth}
        \caption{Zaxxon}
        \includegraphics[width=\linewidth]{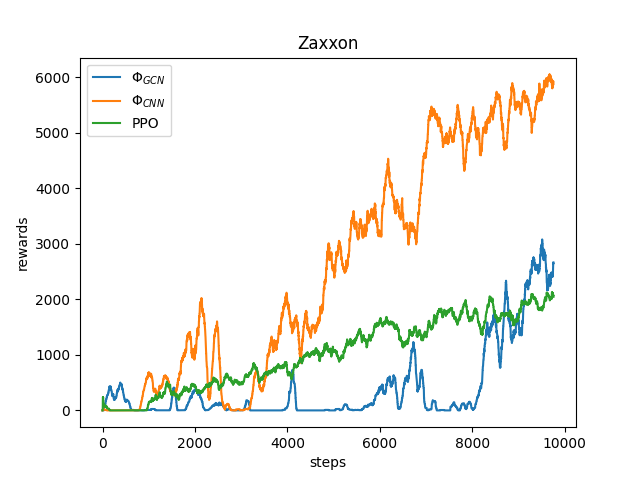}
    \end{subfigure}
    
    \caption{Results on 20 Atari games comapring the performance of $\phi_{CNN}$ to $\phi_{GCN}$ and PPO}
    \label{fig:res_atari_det}
\end{figure*}

These results highlight the capabilities of performing value iteration or message passing using the proposed CNN architecture. Compared to $\phi_{CNN}$, LIRPG does not guarantee invariance for the optimal policy, thus it is not potential-based. On the other hand, RND and ICM provide reward shaping through exploration and can only support discrete action spaces. Additional results showing the performance comparison between $\phi_{CNN}$, $\phi_{GCN}$, and PPO are presented in Figure \ref{fig:res_atari_det}.\\

\subsubsection{\textbf{Mujoco}}
\label{sec:mujoco}
\begin{figure*}[!htb]
    \begin{center}
        \begin{subfigure}[b]{0.49\textwidth}
        \includegraphics[width=\linewidth]{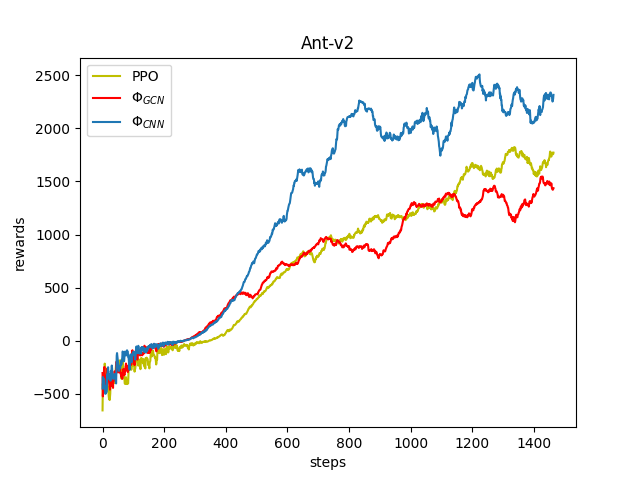}
        \caption{Ant}
        \label{rfidtest1_xaxis}
    \end{subfigure}
    \begin{subfigure}[b]{0.49\textwidth}
        \includegraphics[width=\linewidth]{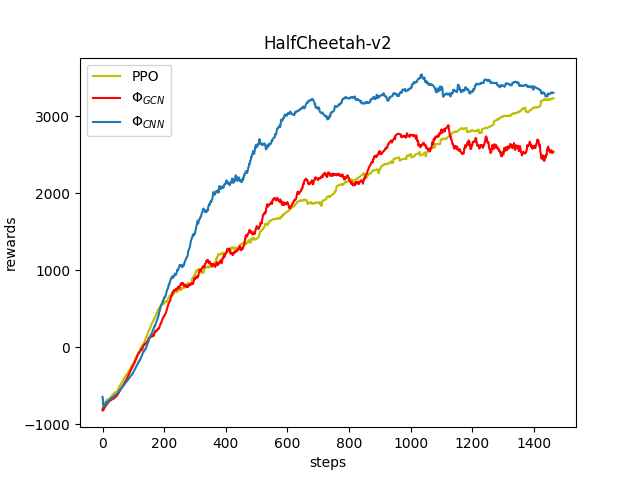}
        \caption{HalfCheetah}
        \label{rfidtest2_xaxis}
    \end{subfigure}
    \end{center}
    
    \begin{center}
        \begin{subfigure}[b]{0.49\textwidth}
        \includegraphics[width=\linewidth]{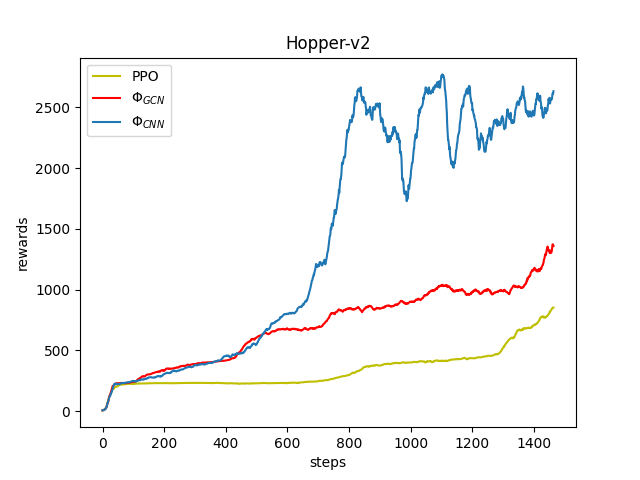}
        \caption{Hopper}
        \label{rfidtest3_xaxis}
    \end{subfigure}
    \begin{subfigure}[b]{0.49\textwidth}
        \includegraphics[width=\linewidth]{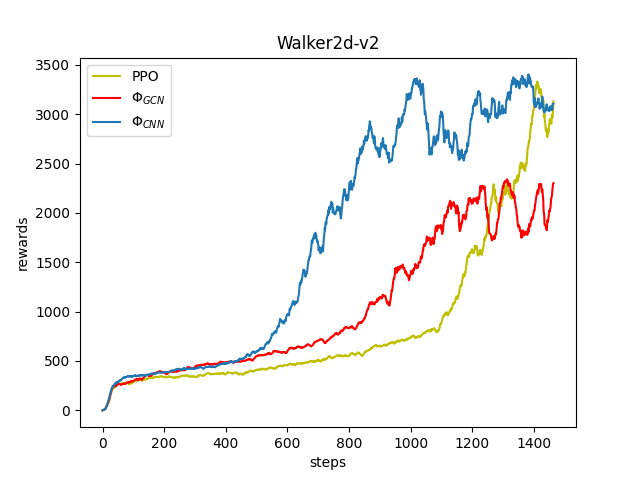}
        \caption{Walker2d}
        \label{rfidtest4_xaxis}
    \end{subfigure}
    \end{center}
    \caption{Performance comparison between different reward shaping mechanisms and PPO in Mujoco environments.}
    \label{fig:res_mujoco}
\end{figure*}

We evaluate the performance of VIN-RS in continuous action space on the MuJoco games compared to baselines. In the CNN implementation of VIN-RS, we deal with the continuous action space by discretizing the action space. The values of two actions at the output layer resemble the range of the continuous actions in $\mathcal{M}$. The rest of the implementation is similar to the Atari games. In terms of baselines comparison, we compare with PPO and $\phi_{GCN}$. It is not possible to compare with RND and ICM because these solutions do not support continuous control. We only compare with $\phi_{GCN}$ as a reward shaping solution because it is the closest to our work in terms of potential-based solution and the use of message passing. The results comparing the performance of each game to the other baselines are shown in Figure \ref{fig:res_mujoco}. The different settings related to this experiment are provided in Table \ref{tab:mujoco_param}. We use the same parameters as in \cite{klissarov2020reward} for evaluating the different techniques and baselines for fair comparison. The experiments were executed for three million steps for each game. We performed the experiment for each solution on each game for five times. The results in Figure \ref{fig:res_mujoco} show the mean for each of the games in terms of rewards with respect to the number of steps in the environment.
\begin{table}[H]
    \centering
    \caption{VIN-RS and RL configurations for the MuJoCo games}
    \begin{tabular}{c|c}
        Hyperparameter & Value \\
        \hline
        Learning rate & 3e-4\\
        $\gamma$ & 0.99\\
        $\lambda$ & 0.95\\
        Entropy Coefficient & 0.0\\
        PPO steps & 2048\\
        PPO Clipping Value & 0.1\\
        \# of minibatches & 32\\
        \# of processes & 1\\
        CNN/GCN (Walker and Ant): $\alpha$ & 0.6\\
        CNN/GCN (Hopper and HalfCheetah): $\alpha$ & 0.6\\
        CNN/GCN: $\eta$ & 1e1\\
      
    \end{tabular}
    \label{tab:mujoco_param}
\end{table}
As shown in the results of Figure \ref{fig:res_mujoco}, the proposed $\phi_{CNN}$ solution using VIN-RS outperforms PPO and $\phi_{GCN}$ in all the MuJoCo games. This improvement is measured by the convergence speed and the ability to reach high rewards at early stages of learning. In addition, the planning capability of $\phi_{CNN}$ is reflected in the performance by reaching high rewards that are not observed by the other solutions. In the results of the Hopper game of Figure \ref{rfidtest3_xaxis}, $\phi_{CNN}$ reaches higher rewards, which is at least two times better than $\phi_{GCN}$ and PPO.

\section{Conclusion and Discussion}
\label{sec:conclusion}

In this paper, we propose VIN-RS, a potential-based reward shaping mechanism that employs a novel CNN architecture as a potential function. VIN-RS performs planning in the environment by implementing a value iteration functionality. The loss function of CNN for value iteration is computed using the message passing mechanism that embeds forward and backward messages, resulting in an effective potential function for rewards shaping. The training of CNN is performed on a sample of transitions. Our solution embeds the look-ahead advice mechanism inside the design of CNN for VIN-RS. We then propose the use of CNN as the potential function to produce shaping values. The resulting shaping values from sampled trajectories at each action are passed to the RL algorithm to update the policy. In terms of learning speed, VIN-RS achieves state-of-the-art results in various games. Besides, through planning, the CNN for reward shaping can converge faster and reach high rewards that are not observable by other solutions. In our evaluations, we showed that the complexity of combining VIN-RS with other RL solutions is minimal. Furthermore, we evaluated the performance of VIN-RS compared to other baselines in Tabular, Atari 2600, and MuJoCo environments.

In terms of future work, the current implementation of VIN-RS does not support high-dimensional state spaces, such as using the MiniWorld environment. This issue can be solved following the VIN generalization solution proposed in \cite{niu2018generalized}.

Using VIN-RS, we believe that existing RL-based solutions in the studied MDP environments have the potential to improve the learning speed and quality of the decisions at the same time. For example, there are various dynamically changing and time-sensitive MDPs requiring fast adaptation to environmental changes. Examples of these environments include autonomous driving, resource management \cite{sami2021demand, sami2021ai}, health applications, Blockchain \cite{kadadha2022chain, hammoud2020ai}, and financial applications \cite{tsantekidis2020price}.

\bibliography{bibfile}

\end{document}